\documentclass{article}

\usepackage[dblblindworkshop, final]{neurips_2026}
\workshoptitle{Foundation Models for Temporal Systems}

\usepackage[utf8]{inputenc}
\usepackage{graphicx}
\usepackage{amsmath}
\usepackage{booktabs}
\usepackage{placeins}
\usepackage[hidelinks]{hyperref}

\newcommand{\boldfigcaption}[2]{\caption{#1}\label{#2}}

\title{Checkpoint Reload Explains Most of the Measured State Gap in Two
State Discovery Reproductions}

\author{
  Weijia Han\thanks{Equal contribution.} \\
  University of Washington
  \And
  Lisha Qu\footnotemark[1] \\
  University of Washington
}

\hypersetup{
  pdftitle={Checkpoint Reload Explains Most of the Measured State Gap in Two State Discovery Reproductions},
  pdfauthor={Weijia Han and Lisha Qu}
}

\begin{document}
\maketitle

\begin{abstract}
Temporal models that maintain per-sequence latent states during training
must reconstruct those states when the stored table is unavailable at
inference. This creates a basic consistency question: do the reconstructed
states agree with those formed during training? We study this question for
State Discovery in World Machine, which updates stored states in parallel
each epoch but performs inference by sequential rollout from an all-zero
start. On Toy1D, we analyze two paired-checkpoint runs whose decoded-state MSE
is 27 to 36 percent above the published value. We compare the stored table with rollouts under the
best-validation checkpoint reloaded after training and the last-epoch
weights that wrote the table. Matching the rollout to the
table-writing weights removes 90 to 91 percent of the mean per-dimension
Wasserstein-1 distance. Causal distillation and frozen SD updates do not
satisfy their pre-specified criteria under either checkpoint. Overall,
checkpoint reload accounts for most of the measured gap in these runs.
\end{abstract}

\section{Introduction}
\label{sec:intro}

Latent world models compress observation histories into states used for
prediction and control, including recurrent state-space models
\citep{hafner2019planet,hafner2023dreamerv3} and recent robust latent control
systems \citep{hansen2024tdmpc2,sun2024robust}. Related work connects state
and history through self-predictive abstractions \citep{ni2024bridging} and
regularizes latent dynamics to make action effects more systematic
\citep{saanum2024simplifying}. Recent studies examine latent-variable recovery
\citep{zhang2025when}, context inferred from interaction \citep{roder2025dali},
and horizon-aware or in-context dynamics
\citep{wan2026uncertain,wang2026context}. These advances make the
state-construction procedure itself consequential. When training stores a
state for each sequence but inference must reconstruct it from a cold start,
evaluating the learned representation requires knowing whether the two states
are consistent. World Machine
\citep{nascimento2026worldmachine} is a concrete instance. Its State
Discovery (SD) procedure writes the core's output back to a per-sequence
state table after each epoch. At inference, the core instead runs
sequentially from an all-zero start. The original paper notes the repeated
updates and all-zero start but does not name or measure this gap.

Recurrent experience replay provides a close analogue: parameter lag can
make stored recurrent states stale, while burn-in recomputes them under
newer weights \citep{kapturowski2019recurrent}. Our setting differs because
the state table is written during validation and the mismatch appears after
early stopping reloads a different checkpoint.

We ask whether SD guarantees agreement (Section~\ref{sec:hardem}), how much
of the gap remains under a shared checkpoint (Section~\ref{sec:gap}), and whether
causal distillation or frozen refinement can reduce the residual without the
table (Sections~\ref{sec:distill} and \ref{sec:fixedpoint}). Checkpoint reload
accounts for nine tenths of the measured gap, while neither procedure meets
its pre-specified success criteria.

\section{The State Discovery Update}
\label{sec:hardem}

\textbf{Notation.} Each sequence carries a per-timestep latent state
$w_t \in [-1,1]^d$ under the model's tanh bound, observations $o$, and a
shared core $F_\theta$. Stored state updating and cold-start inference are
\[
\begin{aligned}
  \Phi_\theta(w) &= \mathrm{shift}\!\left(F_\theta(w,o)\right), \\
  w &\leftarrow \Phi_\theta(w),
      & w_0&\text{ remains fixed}, \\
  \widehat w_0 &= 0,
      & \widehat w_{t+1}&=F_\theta(\widehat w,o)_t.
\end{aligned}
\]
The shift moves the core's per-timestep output forward by one position.
Position 0 is replaced with the input state rather than blended with it,
which is the REPLACE convention. The first two lines constitute SD's
``E-step'' as we read it. They perform fixed-point iteration without a
likelihood comparison, candidate scoring, or an explicit $\arg\max$.

\textbf{Observation 1.} A hard assignment can be viewed as the $T\to0$
limit of a temperature-scaled soft assignment concentrating on
$\arg\max_w p_\theta(o,w)$
\citep{dempster1977em,celeux1992classification}. SD evaluates no
$p_\theta(o,w)$ at candidate states and compares no candidates; it therefore
should not be interpreted as an $\arg\max$. EM's monotone-improvement
guarantee does not transfer: its fixed points need not maximize any
$p_\theta(o,w)$. It is per-sample latent-state iteration without an encoder or
cross-sequence prior, related in form to direct latent-code optimization schemes
\citep{bojanowski2018glo,park2019deepsdf}.

\section{Measuring the Train-Inference Gap Under Both Checkpoints}
\label{sec:gap}

Toy1D is the upstream synthetic benchmark. We train World Machine with SD
enabled and compare the stored and cold-start state populations on the same
8000 held-out items. The stored population is the validation state table,
updated each epoch without gradient flow. The cold-start population is a
fresh sequential rollout of the trained model. We report pooled and
position-resolved 1-Wasserstein distance, denoted $W_1$, together with a
held-out linear probe. Appendix~\ref{app:repro} gives the full benchmark and
split construction; Table~\ref{tab:training_data} lists the configuration.

Unless stated otherwise, population summaries use all 8000 validation
items. Table~\ref{tab:main} and the procedures in Sections~\ref{sec:distill}
and \ref{sec:fixedpoint} use a fixed 1600-item analysis subset. In
Table~\ref{tab:main}, transient $W_1$ is the maximum over positions 1--15;
the receptive-field sweep instead uses positions 1 through the corresponding
receptive field. Means over positions 30--199 are descriptive late-position
summaries.

\textbf{Two checkpoints.} The upstream early stopper reloads the best
validation checkpoint after the final validation pass has written the stored
table. The resulting comparison has three state populations:

\begin{center}
\small
\begin{tabular}{@{}p{0.23\linewidth}p{0.51\linewidth}p{0.16\linewidth}@{}}
\toprule
Population & Construction & Checkpoint \\
\midrule
Stored table & Parallel SD update during the final validation pass
             & $\theta_{\mathrm{last}}$ \\
Reloaded rollout & Sequential rollout from zero after checkpoint reload
                 & $\theta_{\mathrm{best}}$ \\
Matched rollout & Sequential rollout from zero using weights saved before reload
                & $\theta_{\mathrm{last}}$ \\
\bottomrule
\end{tabular}
\end{center}

The trainer normally discards $\theta_{\mathrm{last}}$, so we save it before
reload and roll the cold start under both checkpoints
(Appendix~\ref{app:repro}). With a causal core and position 0 pinned to zero,
the table and rollout should agree at position 1 when they share one checkpoint.
They agree to numerical precision under $\theta_{\mathrm{last}}$ but not
under $\theta_{\mathrm{best}}$ (Figure~\ref{fig:gaprf}).

\textbf{Reproduction context.} The analyzed runs do not match the published
error, as quantified in Section~\ref{sec:discussion}. A separate reproduction
run supplies the receptive-field sweep. Table~\ref{tab:reproduction_runs}
reports all four runs.

\textbf{Checkpoint reload creates most of the gap.} Under
$\theta_{\mathrm{best}}$, the position-resolved gap is largest near the
start and then decays (Figure~\ref{fig:gaprf}). Under
$\theta_{\mathrm{last}}$, the same table is about one tenth as far from the
rollout. Matching the checkpoint removes 91 and 90 percent of the gap for the two
seeds, respectively. Item by item, RMS error falls from $0.36$ to $0.14$ for
seed 42 and from $0.43$ to $0.16$ for seed 7.
Table~\ref{tab:main} gives the analysis-split comparison, and
Table~\ref{tab:paired_comparisons} reports the complete values.

\textbf{Probe safeguard.} On the item-level split, stored and cold-start states
both permit linear recovery of the decoded coordinate. Because sibling windows
can span the split, we treat this as a permissive safeguard, not evidence of
cross-sequence generalization (Appendix~\ref{app:repro}).

\begin{figure}[t]
  \centering
  \includegraphics[width=0.68\textwidth]{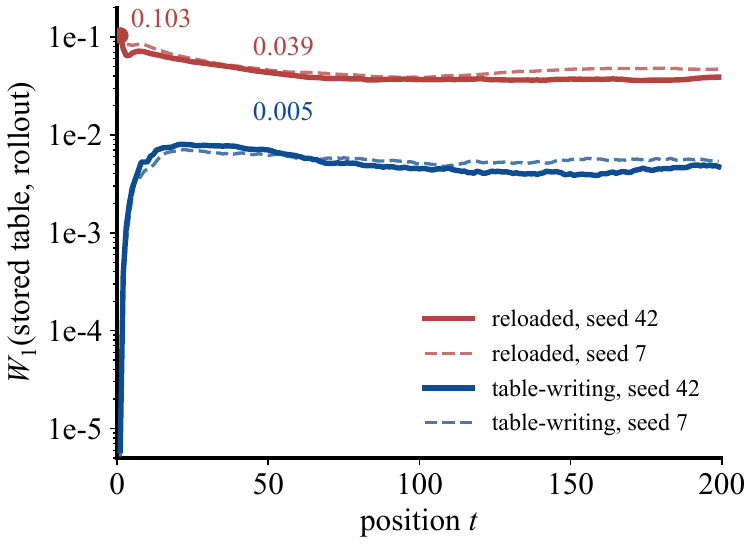}
  \boldfigcaption{Per-position $W_1$, stored table versus cold-start rollout (log
  axis, positions 1 to 199), seed 42 solid, seed 7 dashed, under
  $\theta_{\mathrm{best}}$ (red) and $\theta_{\mathrm{last}}$ (blue);
  labels: the seed-42 position-1 value and the seed-42 means over positions
  30 to 199.}{fig:gaprf}
\end{figure}

\section{Distillation into a Causal TCN Does Not Remove the Gap}
\label{sec:distill}

To test whether the residual is an ordinary amortization gap
\citep{cremer2018amortization}, we train a causal TCN encoder
\citep{bai2018tcn} that maps causal observation windows to stored SD states.
An exploratory four-RF sweep on the first seed-42 run is mixed-checkpoint: its
table uses $\theta_{\mathrm{last}}$, whereas its rollout baseline uses reloaded
$\theta_{\mathrm{best}}$. We use it only for receptive-field trends;
checkpoint-controlled evidence comes from the RF-63 student in the two paired
runs. Success requires cuts of at least
50 percent in both mean and transient $W_1$, with probe $R^2$ within 0.01.

\textbf{None meets the success criterion.} In the exploratory sweep, mean and
transient $W_1$ are larger than the rollout baseline at every RF. Pointwise
fitting error falls with wider context, but mean $W_1$ remains at least 2.9
times the rollout gap at its best RF (Figure~\ref{fig:rfsweep}). In the paired
RF-63 comparisons, the mean-$W_1$ cut is negative for both seeds under both
checkpoints, and the student is also farther from the table item by item
(Table~\ref{tab:main}; Appendix~\ref{app:distill}).

\section{Frozen SD Updates Do Not Meet the Success Criterion}
\label{sec:fixedpoint}

Because distillation does not close the gap, we next test iterative
refinement \citep{kim2018semiamortized,marino2018iterative} by applying $K$
frozen parallel SD updates using $\Phi_\theta$ from
Section~\ref{sec:hardem}. We use the same analysis split and four
initialization types: random states, the distilled student, cold start, and
the stored table. The pre-specified comparison applies five updates to the
student under each checkpoint. Appendix~\ref{app:fixedpoint} lists all pre-specified and
exploratory update counts.

\textbf{Cold start is a fixed point by construction.} For the sequential
rollout $\widehat w$, causality gives $\Phi_\theta(\widehat w)=\widehat w$
analytically, and the implementation agrees to six decimals. More generally,
from any initialization with $w_0=0$, $K$ updates force positions 1 through
$K$ to their cold-start values; refinement is measured on the remaining
positions (Appendix~\ref{app:fixedpoint}).

\textbf{The student fails after five updates under both checkpoints.} Under
$\theta_{\mathrm{best}}$, it produces only a modest,
seed-dependent change and misses the 50 percent threshold. Under
$\theta_{\mathrm{last}}$, its marginal $W_1$ remains 2.5 to 4.7 times the
rollout residual instead of falling below half. Its transient-$W_1$ cut also
falls short, as do all other student update counts
(Table~\ref{tab:main}, Appendix~\ref{app:fixedpoint}).

\textbf{Stored-table initialization.} The stored table is not a fixed point
under either checkpoint. Under $\theta_{\mathrm{last}}$, the $K{=}1$ stored row
is the only row that clears every threshold for both seeds; under
$\theta_{\mathrm{best}}$, it does not. Because this control begins from the
scored table, it is not an inference procedure, and the criterion applies to
student rows. By $K{=}10$, the reloaded-checkpoint trajectories
approach $0.03$--$0.05$ from the table; no comparable floor appears under the
table-writing checkpoint over the evaluated update counts
(Figure~\ref{fig:fixedpoint}, Appendix~\ref{app:fixedpoint}).

\begin{table}[t]
\centering
\caption{Main comparison on the fixed 1600-item analysis subset. Paired
entries give seed 42 followed by seed 7 in parentheses. Mean $W_1$ and
transient $W_1$ (maximum over positions 1--15) measure distance from the
stored table, so lower values are better. Cut is the reduction in mean
$W_1$ relative to the rollout under the same checkpoint.}
\label{tab:main}
\setlength{\tabcolsep}{2pt}
\footnotesize
\begin{tabular}{llcccc}
\toprule
Condition & Checkpoint & Mean W1 & Cut (\%) & Transient W1 & Probe R\textsuperscript{2} \\
\midrule
Cold-start rollout   & $\theta_{\mathrm{best}}$ & 0.03440 (0.03768) & --                  & 0.10368 (0.11391) & 0.9038 (0.8758) \\
Cold-start rollout   & $\theta_{\mathrm{last}}$ & 0.00309 (0.00325) & --                  & 0.00819 (0.00724) & 0.9075 (0.8777) \\
TCN, RF 63           & $\theta_{\mathrm{best}}$ & 0.09046 (0.12383) & $-163.0$ ($-228.6$) & 0.32190 (0.37755) & 0.9012 (0.8543) \\
TCN, RF 63           & $\theta_{\mathrm{last}}$ & 0.09046 (0.12383) & $-2829.6$ ($-3704.4$) & 0.32190 (0.37755) & 0.9012 (0.8543) \\
TCN + 5 SD updates   & $\theta_{\mathrm{best}}$ & 0.02975 (0.03868) & $+13.5$ ($-2.7$)    & 0.10368 (0.11391) & 0.9070 (0.8745) \\
TCN + 5 SD updates   & $\theta_{\mathrm{last}}$ & 0.00761 (0.01542) & $-146.4$ ($-373.6$) & 0.03596 (0.03853) & 0.9095 (0.8747) \\
\bottomrule
\end{tabular}
\vspace{2pt}

\parbox{0.96\linewidth}{\small Success requires 50 percent cuts in both $W_1$
measures and probe R\textsuperscript{2} within 0.01 of stored-state baselines
(0.9076, 0.8762). Each probe uses its row's states. One RF-63 student is scored
against both checkpoints; transient detail is in Appendix~\ref{app:fixedpoint}.}
\end{table}

\begin{figure}[!htb]
  \centering
  \includegraphics[width=0.80\textwidth]{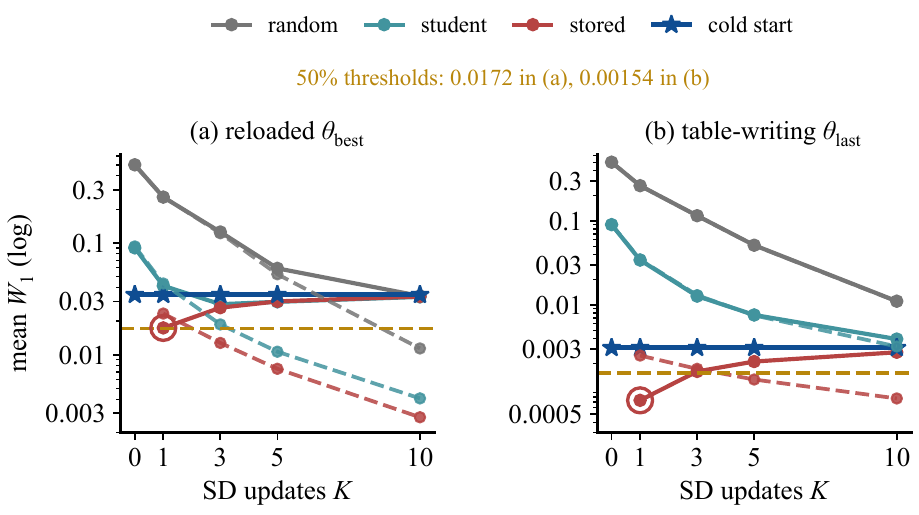}
  \boldfigcaption{Frozen SD-update trajectories for seed 42 under (a) reloaded
  $\theta_{\mathrm{best}}$ and (b) table-writing $\theta_{\mathrm{last}}$.
  Solid: from the table; dashed: from cold start; gold: 50 percent threshold.
  By $K{=}10$, (a) approaches $0.03$; no comparable floor is observed in (b)
  through $K{=}10$ (seed 7 and definitions: Appendix~\ref{app:fixedpoint}).}{fig:fixedpoint}
\end{figure}

\section{Discussion and Limitations}
\label{sec:discussion}

The shared-checkpoint residual exceeds the numerical floor by more than fifty
thousandfold, but its mechanism remains unknown. Among table-free methods, the
$\theta_{\mathrm{last}}$ rollout is closest; refinement does not improve both
measures at once. Matching the rollout to the table-writing checkpoint removes
nine tenths of the gap here, but the size of this effect elsewhere is unknown.

\textbf{Limitations.} The paired analysis covers two seeds on one benchmark;
their decoded-state MSE is 36 and 27 percent above the published value, so
checkpoint separation at that value remains unknown. The RF sweep uses one
seed-42 run without a saved $\theta_{\mathrm{last}}$; its mixed-checkpoint
baseline eases the success threshold, although the RF-63 student fails under
both checkpoints. Marginal $W_1$ ignores item pairing; Appendix~\ref{app:repro}
reports paired errors but not their distributions or downstream costs. No
eligible student met the pre-specified success criterion.

\clearpage
\bibliography{refs}

\appendix
\section{Experimental and Evaluation Details}
\label{app:repro}
This appendix separates the experimental setting, reproduction quality,
checkpoint construction, and evaluation measures. Seed 42 is reported
before seed 7 throughout.

\subsection{Training and data}

The configuration follows the published setup and retains the reproduction's
state-save behavior. The upstream \emph{ShortTimeRecaller} is an auxiliary
short-horizon decoder whose parameters use the fixed base learning rate rather
than the cosine schedule, matching the published training behavior.

\begin{table}[htb]
\centering
\caption{Toy1D training and data configuration.}
\label{tab:training_data}
\small
\begin{tabular}{@{}p{0.28\linewidth}p{0.64\linewidth}@{}}
\toprule
Setting & Specification \\
\midrule
Model and state & 128-dimensional states; 200-position context; two segments; four attention heads; ALiBi positional encoding; tanh activation; REPLACE state saving \\
Optimizer & AdamW with weight decay $5\times10^{-5}$ \\
Training & 1000 epochs with batch size 256 \\
Learning rate & Base rate $10^{-3}$ with cosine warm restarts every 25 optimizer steps; the schedule advances once per batch \\
Toy1D generation & 10000 sequences of 1000 steps from a damped linear system with a 3-dimensional latent state \\
Observations & Random 2-channel linear readout with noise standard deviation 0.1; the objective decodes latent dimension 0 \\
Data split & First 60\% of sequences for training, next 20\% for validation, and final 20\% for testing; 2000 validation sequences with four 200-step windows give 8000 validation items \\
\bottomrule
\end{tabular}
\end{table}

Settings not listed in Table~\ref{tab:training_data} are inherited unchanged
from the published Toy1D configuration. The exploratory RF sweep ran on an
NVIDIA RTX A5000 with internal implementation commit \texttt{a039ac8}. The
paired runs used the same Slurm GPU node and PyTorch 2.13.0+cu126 at internal
implementation commit \texttt{b10bb81}. The available experiment log reports
CUDA compatibility 12.4 but does not preserve the exact NVIDIA driver version
or GPU model. All reported state populations use the validation table;
training-split states are not measured.

\subsection{Reproduction quality}

The pre-specified reproduction band allowed a 20 percent deviation from the
published decoded-state MSE of $0.34348$. Table~\ref{tab:reproduction_runs}
reports the four completed runs.

The August 20 log specified the 20 percent reproduction band, and the August 21
distillation protocol specified the 50 percent mean- and transient-$W_1$ cuts
and 0.01 probe slack. The pre-specified August 24 plan retained these thresholds
for frozen updates and designated the $K{=}5$ student as primary. The September
2 entry specified the paired-checkpoint analysis order, validity checks, and
decision rule before inspection of the corresponding outputs.

Both seed 42 repetitions lie outside the band. Seed 7 lies inside it in the
first repetition and outside it in the second. The result therefore varies
across repetitions as well as seeds, and four runs cannot characterize the
two sources of variation separately. Because the two seed 7 repetitions used
different PyTorch/CUDA builds, we cannot separate stochastic variation from
software-version effects. All four training runs completed 1000 epochs without
numerical failure and produced complete evaluation records. The two
paired-analysis runs used the same compute environment and ran concurrently for
13 hours each. They were designated for paired-checkpoint analysis because
they were instrumented to preserve $\theta_{\mathrm{last}}$; this choice
preceded inspection of their outcomes. The initial strict-load conversion of
both pre-reload snapshots failed because each retained 40 training-only
auxiliary decoder tensors. We repeated only conversion and downstream
evaluation from the unchanged snapshots; no training or parameter updates were
repeated.

Across the four runs, decoded-state MSE is 2.0 to 35.7 percent above the
published value. The paired-analysis runs miss it by 35.7 and 27.3 percent.

\begin{table}[htb]
\centering
\caption{Completed reproduction runs and their use in the paired-checkpoint
analysis.}
\label{tab:reproduction_runs}
\small
\begin{tabular}{@{}cccc@{}}
\toprule
Seed & Repetition & Decoded-state MSE & Used in paired analysis \\
\midrule
42 & first  & 0.46419 & no \\
42 & second & 0.46621 & yes \\
7  & first  & 0.35051 & no \\
7  & second & 0.43723 & yes \\
\bottomrule
\end{tabular}
\end{table}

\subsection{Checkpoints and numerical verification}

The final validation pass writes the stored table under
$\theta_{\mathrm{last}}$. Early stopping then reloads
$\theta_{\mathrm{best}}$. For this analysis, $\theta_{\mathrm{last}}$ was
preserved immediately before the reload and used for a second cold-start
rollout. The saved $\theta_{\mathrm{last}}$ snapshots also contained
training-only auxiliary heads. Conversion removed only those 40 tensors when
reconstructing the inference models; all remaining parameters were preserved
exactly.

The matched-checkpoint identity was verified before we examined the comparison
results.
At position 1, $W_1$ under $\theta_{\mathrm{last}}$ is $0.0000059$ for seed
42 and $0.0000051$ for seed 7. Under $\theta_{\mathrm{best}}$, the
corresponding values are $0.103433$ and $0.112888$. Across all refinement
updates, every cold-start row remains within $0.00000015$ of its matched
rollout. After one update, the differences are $0.00000005$ and $0.00000006$,
and they increase monotonically with the number of updates.

\subsection{Distribution and paired comparisons}

Table~\ref{tab:paired_comparisons} collects the main distributional and
item-level comparisons.

\begin{table}[htb]
\centering
\caption{Distributional and paired checkpoint comparisons. Pooled mean $W_1$
uses all 8000 validation items; paired MSE and RMS use the fixed 1600-item
analysis subset.}
\label{tab:paired_comparisons}
\small
\begin{tabular}{@{}p{0.38\linewidth}cc@{}}
\toprule
Measure & Seed 42 & Seed 7 \\
\midrule
Pooled mean $W_1$, reloaded $\theta_{\mathrm{best}}$ & 0.03438 & 0.03759 \\
Pooled mean $W_1$, table-writing $\theta_{\mathrm{last}}$ & 0.00313 & 0.00361 \\
Per-item MSE, reloaded $\theta_{\mathrm{best}}$ & 0.12779 & 0.18355 \\
Per-item MSE, table-writing $\theta_{\mathrm{last}}$ & 0.01889 & 0.02635 \\
Per-item RMS, reloaded $\theta_{\mathrm{best}}$ & 0.36 & 0.43 \\
Per-item RMS, table-writing $\theta_{\mathrm{last}}$ & 0.14 & 0.16 \\
\bottomrule
\end{tabular}
\end{table}

Under $\theta_{\mathrm{best}}$, the position-resolved curve is zero at
position 0, reaches $0.10343$ and $0.11289$ at position 1, and averages
$0.039$ and $0.045$ over positions 30 through 199. This window was selected
after the primary analysis and is used only for descriptive reporting. Under
$\theta_{\mathrm{last}}$, the curve
is zero to four decimals at positions 0 and 1. It peaks at $0.00801$ at
position 21 and $0.00706$ at position 22, then averages $0.005$ and $0.006$
over positions 30 through 199. Matching the checkpoint removes 91 and 90 percent
of the pooled distance.

Marginal $W_1$ ignores item pairing, so the paired comparison uses
the corresponding stored and rollout states on the 1600-row split. The
matched MSE values are 15 and 14 percent of the reloaded values and 1.7 and
2.4 percent of the random-initialization values of $1.12$ and $1.11$. The
RMS comparison gives a 62 percent reduction for both seeds.

\subsection{Probe protocol}

The probe is an ordinary least-squares linear map from the 128-dimensional
state to the decoded generative coordinate. It is fit on 80 percent of the 8000 items and
scored on the held-out 20 percent, giving $R^2=0.9095$ for seed 42 and
$0.8818$ for seed 7. Because the split is over items, sibling windows from
one sequence can occur on both sides, which makes this estimate optimistic.
We therefore use the probe as a permissive safeguard rather than as evidence
of cross-sequence generalization.
On the 1600-row split of Sections~\ref{sec:distill} and
\ref{sec:fixedpoint}, the same probe is fit on 1280 rows and scored on 320.
It gives $0.9076$ on stored states for seed 42 and $0.8762$ for seed 7. On
cold-start states, the scores are $0.9038$ and $0.8758$ under
$\theta_{\mathrm{best}}$, and $0.9075$ and $0.8777$ under
$\theta_{\mathrm{last}}$.

Together, these checks establish the evaluation populations and the two
checkpoints used by the main comparisons.

\section{Distillation Sweep and Paired-Checkpoint Results}
\label{app:distill}

This appendix asks whether wider causal context improves recovery of the
stored-state distribution. The first seed 42 sweep is mixed-checkpoint: its
table was written under $\theta_{\mathrm{last}}$, its rollout baseline uses
reloaded $\theta_{\mathrm{best}}$, and it preserves no $\theta_{\mathrm{last}}$
for a matched rollout. Its mean $W_1$ is $0.03170$ on the 1600 analysis rows.

\begin{figure}[htb]
  \centering
  \includegraphics[width=0.54\textwidth]{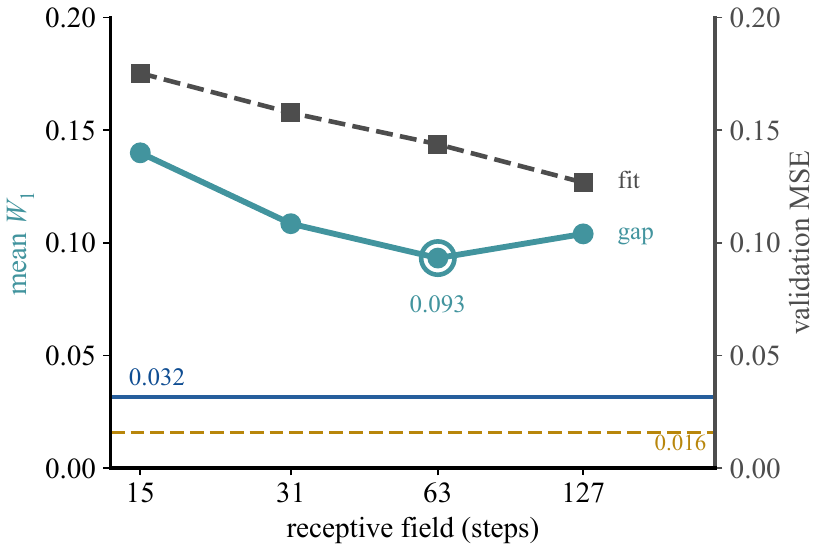}
  \boldfigcaption{Mixed-checkpoint TCN sweep on the first seed-42 run: the table
  uses $\theta_{\mathrm{last}}$ and the rollout baseline uses reloaded
  $\theta_{\mathrm{best}}$. Dashed: validation MSE; solid: mean $W_1$; labels:
  RF 63, rollout gap, and 50 percent threshold.}{fig:rfsweep}
\end{figure}

\subsection{Sweep result}

\begin{table}[htb]
\centering
\caption{Receptive-field sweep on 1600 analysis items (first seed-42 run).}
\label{tab:rf_sweep}
\small
\begin{tabular}{@{}ccccc@{}}
\toprule
Receptive field & Item coverage (\%) & Mean $W_1$ & Cut (\%) & Validation plateau \\
\midrule
15  & 7.5  & 0.13997 & $-341.5$ & 0.175 \\
31  & 15.5 & 0.10851 & $-242.3$ & 0.158 \\
63  & 31.5 & 0.09324 & $-194.1$ & 0.144 \\
127 & 63.5 & 0.10398 & $-228.0$ & 0.127 \\
\bottomrule
\end{tabular}
\end{table}

Item coverage is the receptive field as a percentage of the 200-position item.
Validation plateau is the mean validation MSE over the final ten recorded
epochs. Validation descent is the decrease over the final 20 percent of
training.

Mean $W_1$ is larger than the rollout gap at every receptive field. Transient
$W_1$ is also larger, with cuts from $-147.1$ to $-332.4$ percent.
Validation descent over the final 20 percent of training ranges from 0.0016
to 0.0030. Pointwise fitting error decreases monotonically with wider
context, while mean $W_1$ is lowest at receptive field 63. The difference
between receptive fields 63 and 127 is not interpreted.

Probe $R^2$ improves through receptive field 63 and then decreases slightly at
receptive field 127. For receptive fields 15, 31, 63, and 127, the changes are
$-0.0466$, $-0.0263$, $-0.0095$, and $-0.0132$, respectively; only receptive
field 63 meets the 0.01 slack. Thus the student preserves information about the
decoded coordinate without recovering the stored-state distribution. The probe
has limited discriminating power because its target is close to the decoded
coordinate used by the training objective. At receptive field 63 in the
paired-checkpoint runs, the probe criterion is met for seed 42 but not for seed
7. The complete success criterion is not met for either seed because the mean-
and transient-$W_1$ cuts are negative under both checkpoints.

\subsection{Result on the paired-analysis runs}

For each second repetition, one student is distilled from the stored table
and evaluated under both checkpoints. Its probe $R^2$ change from the
stored-state probe is $-0.0064$ for seed 42 and $-0.0218$ for seed 7. For both
seeds, the transient maximum over positions 1 through 63 occurs within
positions 1 through 15. The values therefore equal those reported in
Table~\ref{tab:main}: $0.32190$ and $0.37755$. The
remaining values appear in Table~\ref{tab:main}.

Item by item, the student's RMS distance from the table is $0.35$ and $0.42$.
The matched-rollout values are $0.14$ and $0.16$, so the student is 2.5 to
2.6 times farther from the table. In marginal $W_1$, it is 29 to 38 times
farther than the matched rollout.

\subsection{Non-comparable RF-190 configuration}

A separate receptive field 190 run is not included in the sweep because it
also changes the kernel size from 3 to 4, confounding the receptive-field
comparison. For transparency, its validation loss was still decreasing at the
end of training: its final-20-percent descent is $0.01472$, 5 to 9 times the
other runs' values. Its final-ten-epoch mean validation MSE is $0.19688$, and
its mean $W_1$ is $0.19170$.

The sweep therefore shows better pointwise fitting with wider context but no
recovery of the stored-state distribution.

\section{Refinement Trajectories by Initialization}
\label{app:fixedpoint}

This appendix asks where repeated frozen SD updates move each
initialization. Results use the second repetitions and 1600 validation rows.
The reported update counts are $K \in \{0,1,3,5,10\}$, with $K{=}10$ treated
as exploratory. Tables~\ref{tab:appc_best} and \ref{tab:appc_last} report
mean $W_1$ to the stored table and to the matched cold-start rollout. The
first random replicate is shown; three random replicates agree within
$0.0004$ at every $K$.

\subsection{Fixed-point identity}

For the sequential rollout $\widehat w$, causality gives
$\Phi_\theta(\widehat w)=\widehat w$ analytically. The numerical implementation
agrees with this identity to six decimal places at every $K$ under both
checkpoints and both seeds. From any initialization with $w_0=0$, $K$ updates
rewrite positions 1 through $K$ with
the cold-start values. The measured rate therefore concerns the remaining
$199-K$ positions, while position 0 stays pinned.

\subsection{Trajectory interpretation}

Under the reloaded checkpoint, every initialization approaches the matched
rollout. At $K{=}10$, the means across three random replicates are $0.03358$
and $0.04566$ from the stored table. Under both checkpoints and both seeds,
cold-start rows remain within $0.00000015$ of the matched rollout at every
$K$. By $K{=}10$, the reloaded-checkpoint trajectories approach values between
$0.03$ and $0.05$ from the table. No comparable floor appears under
$\theta_{\mathrm{last}}$ over the evaluated update counts; the student reaches
1.3 to 1.9 times the rollout residual at exploratory $K{=}10$.

\subsection{Pre-specified comparison}

At $K{=}5$ under
$\theta_{\mathrm{last}}$, the student remains 2.5 to 4.7 times the rollout
residual in marginal $W_1$ and 1.6 to 1.7 times the residual item by item.
One update from the stored table under $\theta_{\mathrm{best}}$ moves it by 51
and 68 percent of the stored-to-cold-start distance. Under
$\theta_{\mathrm{last}}$, the corresponding shares are 24 and 25 percent of
a distance eleven times smaller. The $K{=}1$ stored-table row under
$\theta_{\mathrm{last}}$ is the only row that clears every threshold, with mean
cuts of 76 and 75 percent, transient cuts of 57 and 54 percent, and probe change
within $0.0001$. It starts from the table against which it is scored, so it is
not an inference procedure; the success criterion was pre-specified for student
rows only.

\subsection{Transient and paired comparisons}

Under the reloaded checkpoint, the transient equals the
rollout's from $K\geq3$: $0.10368$ for seed 42 and $0.11391$ for seed 7. The
rollout maximum is at position 1, which every update rewrites, and positions 2
to 15 fall below it by $K{=}3$. Therefore, the transient-$W_1$ column in
Table~\ref{tab:main} cannot distinguish the $\theta_{\mathrm{best}}$ refinement
row from the rollout baseline. Under the
table-writing checkpoint, the rollout transient is $0.00819$ and $0.00724$, while the
student falls from $0.32190$ to $0.01252$ and from $0.37755$ to $0.01246$
across $K=0$ to $10$.

At $K{=}10$, student versus matched rollout
per-item RMS is $0.16$ versus $0.14$ for seed 42 and $0.21$ versus $0.16$ for
seed 7. After ten reloaded updates, table versus matched rollout RMS is $0.16$
versus $0.14$ and $0.19$ versus $0.16$. Thus the marginal agreement in
Tables~\ref{tab:appc_best} and \ref{tab:appc_last} does not imply paired
agreement.

Together, the trajectories show that repeated updates reduce differences due
to initialization but do not make the student match the stored table under
either checkpoint.

\begin{table}[htb]
\centering
\caption{Frozen E-step trajectories under the reloaded $\theta_{\mathrm{best}}$ on 1600 validation rows. Each panel reports mean $W_1$ to the stored table and to the matched rollout. The result at $K{=}10$ is exploratory.}
\label{tab:appc_best}
\setlength{\tabcolsep}{5pt}
\renewcommand{\arraystretch}{1}
\footnotesize
\textbf{Seed 42}\\[2pt]
\begin{tabular}{llccccc}
\toprule
Initialization & Reference & $K{=}0$ & $K{=}1$ & $K{=}3$ & $K{=}5$ & $K{=}10$ \\
\midrule
random & stored table & 0.50136 & 0.25726 & 0.12551 & 0.05930 & 0.03356 \\
 & matched rollout & 0.49663 & 0.25860 & 0.12230 & 0.05252 & 0.01138 \\
student & stored table & 0.09046 & 0.04140 & 0.02833 & 0.02975 & 0.03292 \\
 & matched rollout & 0.09316 & 0.04338 & 0.01867 & 0.01065 & 0.00407 \\
stored & stored table & -- & 0.01738 & 0.02636 & 0.03003 & 0.03321 \\
 & matched rollout & -- & 0.02336 & 0.01276 & 0.00748 & 0.00276 \\
cold start & stored table & 0.03440 & 0.03440 & 0.03440 & 0.03440 & 0.03440 \\
 & matched rollout & 0.00000 & 0.00000 & 0.00000 & 0.00000 & 0.00000 \\
\bottomrule
\end{tabular}
\par\vspace{6pt}
\textbf{Seed 7}\\[2pt]
\begin{tabular}{llccccc}
\toprule
Initialization & Reference & $K{=}0$ & $K{=}1$ & $K{=}3$ & $K{=}5$ & $K{=}10$ \\
\midrule
random & stored table & 0.46405 & 0.27336 & 0.13337 & 0.08090 & 0.04565 \\
 & matched rollout & 0.46045 & 0.26865 & 0.12028 & 0.06338 & 0.01923 \\
student & stored table & 0.12383 & 0.06375 & 0.04213 & 0.03868 & 0.03792 \\
 & matched rollout & 0.11951 & 0.05582 & 0.02659 & 0.01630 & 0.00607 \\
stored & stored table & -- & 0.02573 & 0.03210 & 0.03442 & 0.03701 \\
 & matched rollout & -- & 0.02359 & 0.01335 & 0.00821 & 0.00318 \\
cold start & stored table & 0.03768 & 0.03768 & 0.03768 & 0.03768 & 0.03768 \\
 & matched rollout & 0.00000 & 0.00000 & 0.00000 & 0.00000 & 0.00000 \\
\bottomrule
\end{tabular}
\end{table}

\begin{table}[htb]
\centering
\caption{Frozen E-step trajectories under the table-writing $\theta_{\mathrm{last}}$ on 1600 validation rows. Each panel reports mean $W_1$ to the stored table and to the matched rollout. The result at $K{=}10$ is exploratory.}
\label{tab:appc_last}
\setlength{\tabcolsep}{5pt}
\renewcommand{\arraystretch}{1}
\footnotesize
\textbf{Seed 42}\\[2pt]
\begin{tabular}{llccccc}
\toprule
Initialization & Reference & $K{=}0$ & $K{=}1$ & $K{=}3$ & $K{=}5$ & $K{=}10$ \\
\midrule
random & stored table & 0.50136 & 0.26321 & 0.11560 & 0.05148 & 0.01109 \\
 & matched rollout & 0.50129 & 0.26321 & 0.11548 & 0.05126 & 0.01100 \\
student & stored table & 0.09046 & 0.03424 & 0.01274 & 0.00761 & 0.00393 \\
 & matched rollout & 0.09072 & 0.03475 & 0.01307 & 0.00752 & 0.00320 \\
stored & stored table & -- & 0.00073 & 0.00162 & 0.00212 & 0.00273 \\
 & matched rollout & -- & 0.00250 & 0.00173 & 0.00129 & 0.00077 \\
cold start & stored table & 0.00309 & 0.00309 & 0.00309 & 0.00309 & 0.00309 \\
 & matched rollout & 0.00000 & 0.00000 & 0.00000 & 0.00000 & 0.00000 \\
\bottomrule
\end{tabular}
\par\vspace{6pt}
\textbf{Seed 7}\\[2pt]
\begin{tabular}{llccccc}
\toprule
Initialization & Reference & $K{=}0$ & $K{=}1$ & $K{=}3$ & $K{=}5$ & $K{=}10$ \\
\midrule
random & stored table & 0.46405 & 0.25355 & 0.11581 & 0.06218 & 0.01775 \\
 & matched rollout & 0.46435 & 0.25378 & 0.11559 & 0.06156 & 0.01695 \\
student & stored table & 0.12383 & 0.05250 & 0.02432 & 0.01542 & 0.00625 \\
 & matched rollout & 0.12420 & 0.05287 & 0.02476 & 0.01578 & 0.00617 \\
stored & stored table & -- & 0.00080 & 0.00178 & 0.00235 & 0.00298 \\
 & matched rollout & -- & 0.00263 & 0.00188 & 0.00147 & 0.00099 \\
cold start & stored table & 0.00325 & 0.00325 & 0.00325 & 0.00325 & 0.00325 \\
 & matched rollout & 0.00000 & 0.00000 & 0.00000 & 0.00000 & 0.00000 \\
\bottomrule
\end{tabular}
\end{table}

\FloatBarrier

\end{document}